\documentclass[journal]{IEEEtran} 
\IEEEoverridecommandlockouts
\usepackage{cite}
\usepackage{amsmath,amssymb,amsfonts}
\usepackage{algorithmic}
\usepackage{graphicx}
\usepackage{textcomp}
\usepackage{xcolor}
\usepackage{color}  
\usepackage{hyperref}  
\hypersetup{ 
  colorlinks=true,  
  linkcolor=blue,  
  filecolor=blue,  
  urlcolor=blue,  
  citecolor=cyan,  
}
\def\BibTeX{{\rm B\kern-.05em{\sc i\kern-.025em b}\kern-.08em
    T\kern-.1667em\lower.7ex\hbox{E}\kern-.125emX}}
\begin{document}

\IEEEoverridecommandlockouts

\title{TrafficGPT: Viewing, Processing and Interacting with Traffic Foundation Models\\
}

\author{%
Siyao Zhang$^{1, \dagger}$, Daocheng Fu$^{3, \dagger}$, Zhao Zhang$^{1,2, \ast}$, Bin Yu$^{1,2}$ and Pinlong Cai$^{3}$ 
\thanks{$^{1}$ Siyao Zhang, Zhao Zhang and Bin Yu are with School of Transportation Science and Engineering (Beihang University), Beijing, China. Email: \{siyaozhang, zhaozhang, yubinyb\}@buaa.edu.cn}
\thanks{$^{2}$ Zhao Zhang and Bin Yu are with Key Laboratory of Intelligent Transportation Technology and System (Ministry of Education), Beijing, China.}
\thanks{$^{3}$ Daocheng Fu and Pinlong Cai are with Shanghai Artificial Intelligence Laboratory, Shanghai, China. Email: \{fudaocheng, caipinlong\}@pjlab.org.cn}
\thanks{$^{\dagger}$ These authors contributed equally to this work.}
\thanks{$^{\ast}$ Corresponding author.}
}


\maketitle

\begin{abstract}
 With the promotion of chatgpt to the public, Large language models indeed showcase remarkable common sense, reasoning, and planning skills, frequently providing insightful guidance. These capabilities hold significant promise for their application in urban traffic management and control. However, LLMs struggle with addressing traffic issues, especially processing numerical data and interacting with simulations, limiting their potential in solving traffic-related challenges. In parallel, specialized traffic foundation models exist but are typically designed for specific tasks with limited input-output interactions. Combining these models with LLMs presents an opportunity to enhance their capacity for tackling complex traffic-related problems and providing insightful suggestions.
    
To bridge this gap, we present TrafficGPT—a fusion of ChatGPT and traffic foundation models. This integration yields the following key enhancements: 1) empowering ChatGPT with the capacity to view, analyze, process traffic data, and provide insightful decision support for urban transportation system management; 2) facilitating the intelligent deconstruction of broad and complex tasks and sequential utilization of traffic foundation models for their gradual completion; 3) aiding human decision-making in traffic control through natural language dialogues; and 4) enabling interactive feedback and solicitation of revised outcomes. By seamlessly intertwining large language model and traffic expertise, TrafficGPT not only advances traffic management but also offers a novel approach to leveraging AI capabilities in this domain. The TrafficGPT demo can be found in \href{https://github.com/lijlansg/TrafficGPT.git}{https://github.com/lijlansg/TrafficGPT.git}.
\end{abstract}


\section{Introduction}

Recent advancements in the fields of artificial intelligence (AI) and natural language processing (NLP) have ushered in a new era of possibilities. Large language models, epitomized by ChatGPT\cite{ouyang2022training}, exhibit remarkable common sense, reasoning, and planning abilities, and often providing insightful suggestions. These qualities hold significant promise for urban traffic management and control.

Nevertheless, it's crucial to acknowledge that despite their impressive language capabilities, large language models (LLMs) inherently lack an in-depth comprehension of traffic-related complexities. This inherent limitation further constrains their capacity to proficiently handle numerical data and interact with simulations, ultimately curtailing their effectiveness in addressing intricate traffic challenges.

Concurrently, the field of traffic management has been fortunate to witness the development of a diverse array of Traffic Foundation Models (TFMs). These models have been meticulously designed to address specific and nuanced traffic-related challenges. While these TFMs excel in their designated tasks, they typically operate within the confines of single-round inputs and outputs. Additionally, due to their sheer number and specialization, identifying and stringing together TFMs for complex tasks presents a formidable challenge for human operators.

This presents a unique and promising opportunity to bridge the gap between the capabilities of large language models (LLMs) and the specialized knowledge embedded within TFMs. By seamlessly integrating these distinct strengths, we can pave the way for innovative solutions that significantly enhance comprehension and problem-solving within the field of traffic management. 

\begin{figure*}[htbp]
    \centerline{\includegraphics[width=0.9\textwidth]{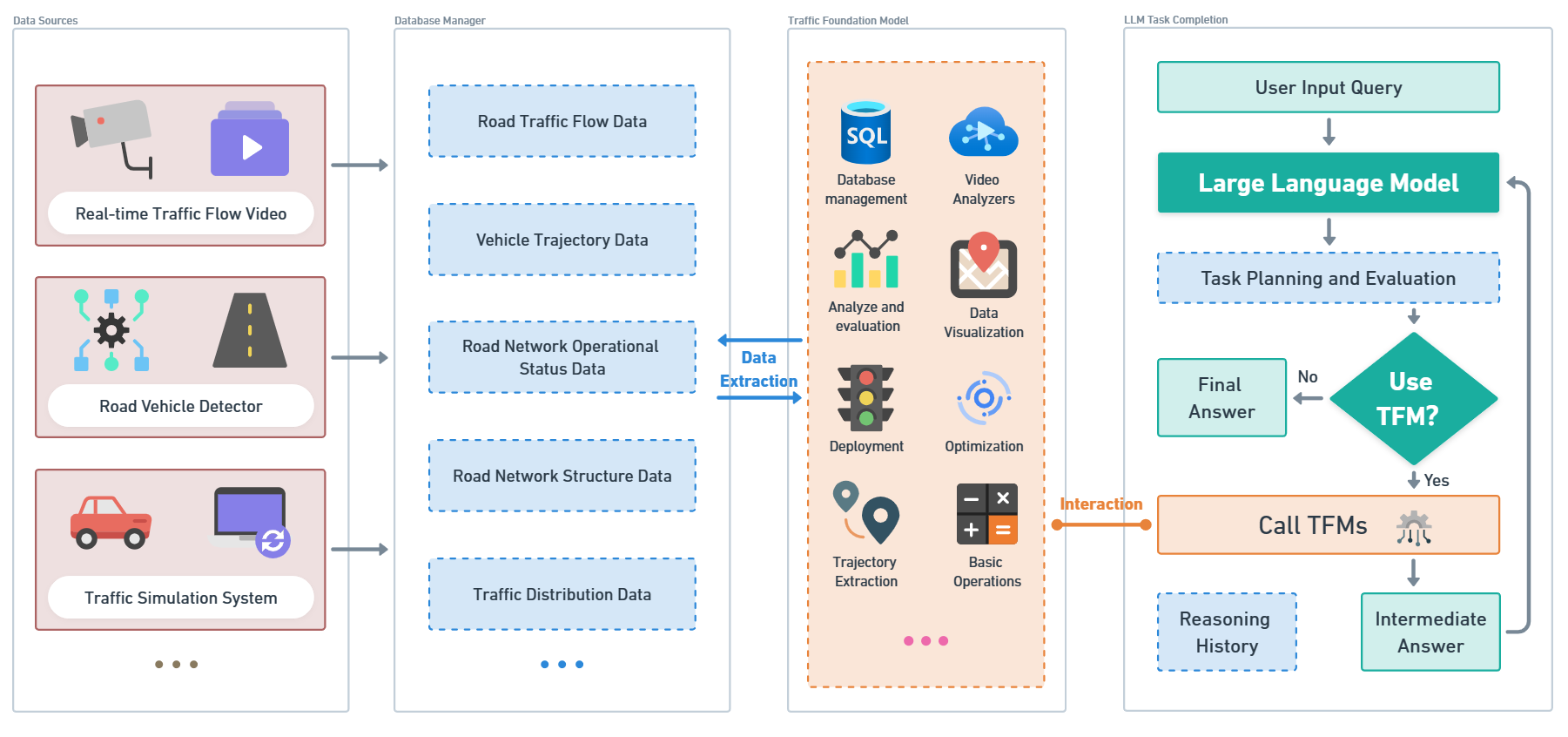}}
    \caption{TrafficGPT Framework: TrafficGPT utilizes multimodal traffic data from sources such as video data, detector data, simulation system data, and others. Instead of direct interaction with these data sources, Traffic Foundation Models (TFM) facilitate data access through an intermediary database manager layer. At the outermost level of the framework, a Large Language Model (LLM) identifies user requirements and orchestrates task execution through TFM.}
    \label{Framework_of_TrafficGPT}
\end{figure*}

Therefore, we propose TrafficGPT, a pioneering amalgamation of ChatGPT and traffic foundation models, to empower LLM with tools to make LLM understand more deeply what it is working on, so that LLM can accomplish some complex operations or give insightful suggestions and decisions to human users. Significantly, TrafficGPT leverages multimodal data as a data source, thereby offering comprehensive support for various traffic-related tasks. The framework of TrafficGPT is shown in ``Fig.~\ref{Framework_of_TrafficGPT}''. This integration holds the promise of revolutionizing traffic management by harnessing AI's potential to tackle the intricate challenges posed by traffic data analysis and decision-making. In the following sections, we will delve into the architecture, methodologies, and outcomes of TrafficGPT, showcasing how it advances the boundaries of AI application in the realm of traffic management.

\section{Related Works}

\subsection{Applications of LLMs in Transportation Field}
With the growing recognition of Large Language Models (LLMs) in diverse domains, And vertical applications that more accurately meet industry-specific needs have been birthed in areas such as education\cite{baidoo2023education,biswas2023role}, finance\cite{zhang2023xuanyuan,xu2023baize}, and vision\cite{wu2023visual,gong2023multimodal}. Scholars in the field of traffic management and control also have made noteworthy contributions, exploring novel approaches and applications.
To test the LLM's knowledge on traffic control, Villarreal et al.\cite{villarreal2023can} investigates the use of ChatGPT, a large language model, to assist novices in solving complex mixed traffic control problems in Intelligent Transportation Systems (ITS). The results of the study show that using ChatGPT can increase the number of successful policies in mixed traffic control tasks.
Further, Da et al.\cite{da2023llm} proposes a PromptGAT framework for traffic signal control, integrating prompt-based dynamics modeling using language models to enhance the learning of forward and inverse models. The framework leverages domain knowledge and real-time traffic states to predict system dynamics and optimize the policy through reinforcement learning.
As for information processing, Zheng et al.\cite{zheng2023chatgpt} explores the potential applications of large language models (LLMs), specifically ChatGPT, in intelligent transportation systems. By leveraging LLMs with cross-modal encoders, they present a smartphone-based crash report auto-generation and analysis framework based on news text, images and audio inputs. The paper discusses the benefits and challenges of using LLMs in transportation, including data privacy, data quality, and model bias. Overall, LLMs hold promise for more efficient, intelligent, and sustainable transportation systems. 

Nonetheless, to the author's best understanding, the incorporation of Large Language Models (LLMs) in complex traffic-related tasks, encompassing analysis, processing, and interaction with expansive traffic-related datasets, remains sparsely explored within the current research milieu. This situation is likely attributed to the inherent challenges associated with the reliability of Large Language Models (LLMs) when handling numerical digits within numeric data\cite{frieder2023mathematical,azaria2022chatgpt}. This dearth of integration underscores a significant gap in the application of LLMs in the field of tackling complex traffic-related issues and providing supported decision making. 

\subsection{Models for Traffic-related Tasks}
Complex traffic-related tasks inevitably involve sub-tasks that requires LLM to deal with numerical traffic datasets. The processing, analysis, and visualization of traffic data are essential to support decision-making among traffic managers. Many widely recognized and used Traffic Foundation Models(TFMs) for traffic-related tasks have emerged from research and practice in the field of transportation.
Traffic system analysis models can be categorized into network topology analysis\cite{kwayu2021discovering,osei2014complex}, network volume analysis\cite{willumsen1978estimation,lingras2000traffic}, and network performance analysis\cite{fisk1991traffic,isradi2020performance,marisamynathan2016performance}.
Apart from pure data analysis using sophisticated models, Chen et al.\cite{chen2015survey} emphasizes the importance of data visualization in understanding the behavior of traffic participants and discovering traffic patterns. Existing data visualization models could present location-based, activity-based, and device-based traffic data \cite{andrienko2008basic,andrienko2012visual} with a variety of focuses including time\cite{guo2011tripvista}, spatial\cite{liu2013vait,xie2008kernel,andrienko2009interactive} and other properties needed \cite{kraak2003space}.
Furthermore, building upon the insights gleaned from traffic data analysis, extensive research and validation efforts have been dedicated to the development of traffic management and control measures. These encompass optimizations in signal schemes\cite{webster1958traffic,varaiya2013max}, route guidance systems\cite{papageorgiou1990dynamic}, and more.

Although these Traffic Foundation Models (TFMs) are often confined to specific tasks with single-round inputs and outputs, the abundance of well-established TFMs, coupled with the prospect of amalgamating multiple TFMs, forms a sturdy basis for deploying Large Language Models (LLMs) to tackle complex traffic-related issues and support decision-making.

\subsection{Empowering LLMs for Complex Traffic-related Tasks}
To deal with complex tasks, it is necessary to uncover the reasoning and planning capabilities of LLMs. The concept of Chain-of-Thought (CoT) \cite{wei2022chain} is introduced to harness the multi-step reasoning capabilities inherent in Large Language Models (LLMs). Further, the ReAct \cite{yao2022react} logic investigate the application of LLMs in generating both reasoning traces and task-specific actions in an intertwined fashion, fostering enhanced synergy between the two components. Reasoning traces serve the purpose of enabling the model to deduce, monitor, and revise action plans, as well as manage exceptions, whereas actions empower it to interface with external resources. However, the application of CoT and ReAct mainly focus on decision making tasks based on text information, i.e., HotpotQA Example \cite{yao2022react}. Our work extends the potentiality of LLM to complex traffic-related tasks regarding expansive numerical datasets, including but not limited to data processing, data visualization, traffic management and control.

\begin{figure*}[htbp]
    \centerline{\includegraphics[width=0.9\textwidth]{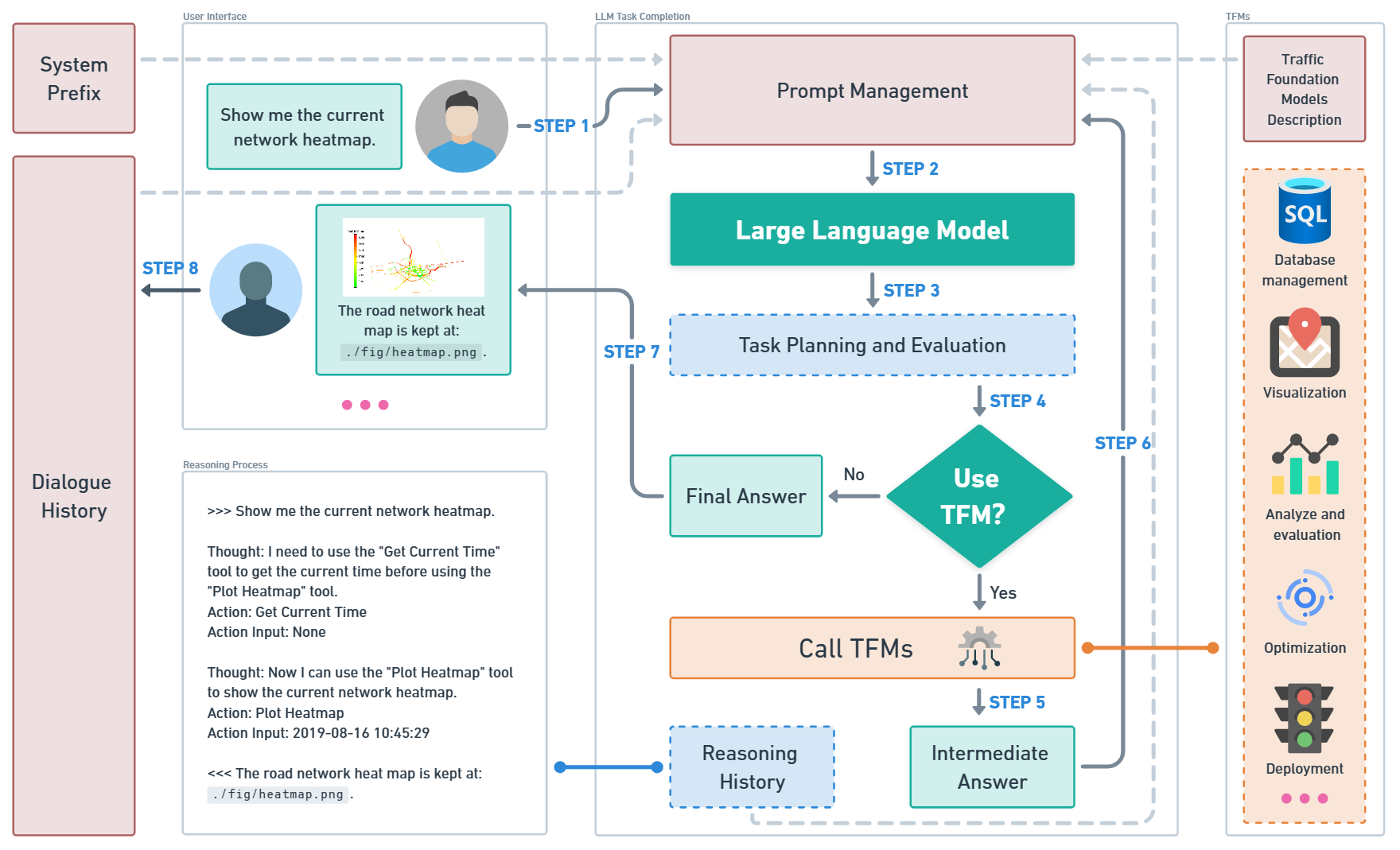}}
    \caption{Overview of TrafficGPT. The User Interface section shows how users interact with TrafficGPT. The Reasoning Process section shows how TrafficGPT deconstruct the task and complete the sub-tasks step by step. The LLM Task Comletion section shows the flowchart of how TrafficGPT form prompts and iteratively invokes Traffic Foundation Models to provide answers.}
    \label{overview TrafficGPT}
\end{figure*}
\section{TrafficGPT}

In this section, we elaborate on the detailed mechanisms of how TrafficGPT utilize large language models to deconstruct and execute complex traffic-related tasks. The proposed process is presented in ``Fig.~\ref{overview TrafficGPT}'' and comprises the following key steps:

-- STEP 1: Natural Language Input: The process begins with users inputting task requirements in natural language through the frontend of TrafficGPT. This input text serves as a prompt and is passed to the next step for prompt management.

-- STEP 2: Prompt Management: As a foundational step, "Prompt Management" is introduced to define the operational framework of the LLM agent. It involves delineating the agent's working mechanism, specifying crucial considerations, and conveying information about the available toolset. Additionally, this step enables the incorporation of historical dialogue context to facilitate multi-turn interactions. The components of this integrated prompt include the User Task Request, a System Prefix, Available Tools, Reasoning History and Dialogue History. By amalgamating these elements into a cohesive prompt, the agent is equipped with the necessary context and instructions to facilitate effective task deconstruction and execution.

-- STEP 3: Natural Language Understanding and Task Planning: Leveraging the capabilities of LLMs, the agent comprehends the prompts in natural language. Owing to the cognitive capabilities inherent in LLMs, the agent undertakes deductive reasoning facilitated by amalgamating the Task Request, the set of Available Tools, and the repository of Reasoning History. This intellectual process culminates in the development of discernible and operational insights referred to as "Thoughts", as presented in the Reasoning Process section of ``Fig.~\ref{overview TrafficGPT}''.

-- STEP 4: Traffic Foundation Model Execution: Drawing from the established Thoughts, the agent calls the selected TFMs among the Available Tools and formulates parameters in strict compliance with the prerequisites delineated in the Tool Definition. Harnessing these parameters, the TFM conducts distinct tasks, encompassing functionalities like database retrieval and analysis, data visualization, and system optimization, culminating in the generation of desired output results.

-- STEP 5: Result Output and Intermediate Answer: Upon the tool's execution, the agent retrieves the output of TFMs through the API interface. The agent integrate the tool's output into Intermediate Answers in the form of natural language for the LLM to do further planning. In scenarios necessitating multi-modal output as supplement information, structured content, such as tables, will be produced in Markdown format, while visualization images, data files, and similar components will be provided in the form of file paths.

-- STEP 6: Task Assessment and Continuation: The agent conducts a comparative analysis between the User Task Request and the ongoing Intermediate Answer to gauge the status of task completion. Should the task remain unresolved, the procedure regresses to STEP 2 through STEP 5, ensuring an iterative continuation of the execution process.

-- STEP 7: Final Answer Generation: After ascertaining the accomplishment of the task in STEP 6, the agent engages the extensive capabilities of LLMs and consolidates the output content generated by the tools to formulate a conclusive response. Subsequently, this meticulously crafted response is transmitted to the user through the frontend interface.

-- STEP 8: Dialogue Memory Storage: In this step, the continuous conversation is preserved by storing both the user inputs and the LLM's outputs. These records are summarized in Dialogue History and serve as a part of the input of Prompt Management in subsequent interactions, providing a conversational context that imbues the large language model with memory capability.

By implementing this comprehensive framework, the amalgamation of large language models with intelligent transportation systems promises to reshape the landscape of traffic data analysis. The ensuing subsections engage in a comprehensive exploration of multiple pivotal elements, providing detailed insights into their inherent significance and key components.

\subsection{System Prefix}\label{AA}
TrafficGPT constitutes a system that synergizes various Traffic Foundation Models (TFMs) to facilitate the comprehension of traffic-related tasks and the subsequent generation of relevant responses based on multi-modal data sources. This integration necessitates the customization of specific system principles, subsequently transformed into prompts comprehensible to Large Language Models (LLMs). These prompts play a multifaceted role, encompassing:
\begin{itemize}
    \item  \textbf{Role of TrafficGPT:} In general, TrafficGPT is designed as an AI to assist human with traffic-related tasks. To further refine the distinction, TrafficGPT can be used for traffic big data analysis and display, supported decision-making of traffic management, and even traffic simulation modeling and control.
    \item  \textbf{TFMs Accessibility:} TrafficGPT is equipped with an array of Traffic Foundation Models (TFMs) to address diverse traffic-related tasks. The selection of the particular foundation model to employ is autonomously determined by the Large Language Model (LLM), thereby facilitating seamless integration of new TFMs and the extension of support to novel traffic-related tasks.
    \item  \textbf{Task Deconstruction and Planning:} Depicted in ``Fig.~\ref{overview TrafficGPT}'', addressing what may initially appear as a straightforward command often entails the involvement of multiple Traffic Foundation Models (TFMs). For instance, consider the query "Show me the current network heatmap," which necessitates the utilization of both the "Get Current Time" and "Plot Heatmap" TFMs. To effectively handle more intricate queries by deconstructing them into subproblems, the introduction of the Thought-Action-Observation mechanism within TrafficGPT becomes essential. This mechanism aids in the determination, utilization, and distribution of multiple TFMs to successfully address the challenges posed by diverse queries.
\end{itemize}

\begin{figure*}[htbp]
    \centerline{\includegraphics[width=0.9\textwidth]{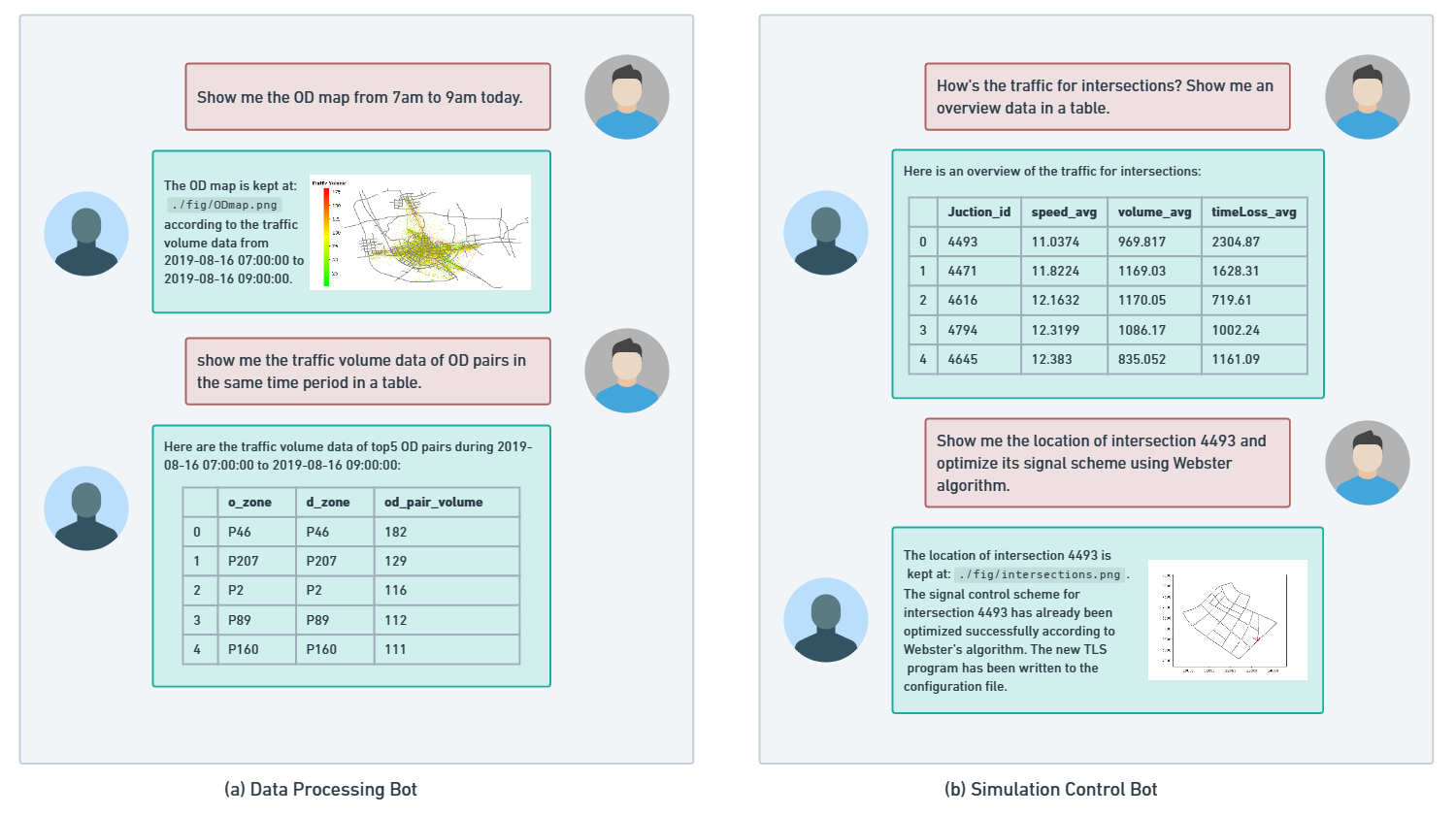}}
    \caption{Basic capability of TrafficGPT to execute traffic-related commands based on natural language comprehension. In the dialogues to the left, Data Processing Bot can access the database and extract traffic data during specific time period, as well as execute complex data visualization command. In the dialogues to the right, Simulation Control Bot demonstrates the ability to perform simulation data retrieving, modeling and control.}
    \label{Example1}
\end{figure*}

\subsection{Traffic Foundation Models}
TrafficGPT has the capacity to incorporate a diverse array of Traffic Foundation Models (TFMs), which encompass an expansive spectrum of functions, encompassing but not limited to database access, traffic flow counting, trajectory extraction, traffic performance evaluation, traffic resilience quantification, data visualization, and traffic signal optimization. These functions are tailored to address the practical requirements of traffic-related tasks.

However, it is noteworthy that distinct TFMs may share similar functionalities. For instance, for signal optimization tasks, TrafficGPT has TFMs based on various methods such as the Webster algorithm, linear programming methods, reinforcement learning techniques, and more. These tools may cater to varied scenarios and possess disparate input requisites.

Consequently, the assurance of the Large Language Model's (LLM) capability to distinctly discern the TFMs and proficiently apply them becomes a crucial factor in facilitating the proficient execution of traffic-related tasks. To attain this objective, we craft prompts and input them into the prompt management system for each TFM. This is accomplished by outlining the subsequent attributes:
\begin{itemize}
    \item \textbf{Name:} The name usually gives an overview of the functionality implemented by each TFM. Its core function is to provide the LLM with an entry point to the TFM. The target TFM can only be called and run if the LLM provides an identical tool name.
    \item \textbf{Usage:} The usage mainly introduces the actual application scenarios and the methodologies employed of each TFM, so that LLM can distinguish between similar TFMs.
    \item \textbf{Input:} The definition of the TFM input content is key to ensuring that TrafficGPT is able to successfully invoke the tool. The input typically describes the content and format of the input parameters required to invoke each TFM. Often, textual descriptions are followed by examples to ensure that the LLM is accurately understood. For instance, the TFM used for plotting heatmap, the Input part of the TFM's Prompt is as follows: The input must be the target point in time in the "YYYY-MM-DD HH:MM:SS" format. For example: 2019-08-13 09:00:00.
    \item \textbf{Output:} The output describes the content and format of the output of each TFM.
    \item \textbf{Priority(Optional):} In order to improve the robustness of user demand response, for similar TFMs, in addition to providing information to help differentiate between them, another possible approach is to set a higher priority for certain TFMs for similar tasks. For example, when a user proposes a signal scheme optimization task but does not explicitly indicate a specific algorithm, the signal optimization TFM based on Webster's algorithm is preferred.
\end{itemize}

\begin{figure*}[htbp]
    \centerline{\includegraphics[width=0.9\textwidth]{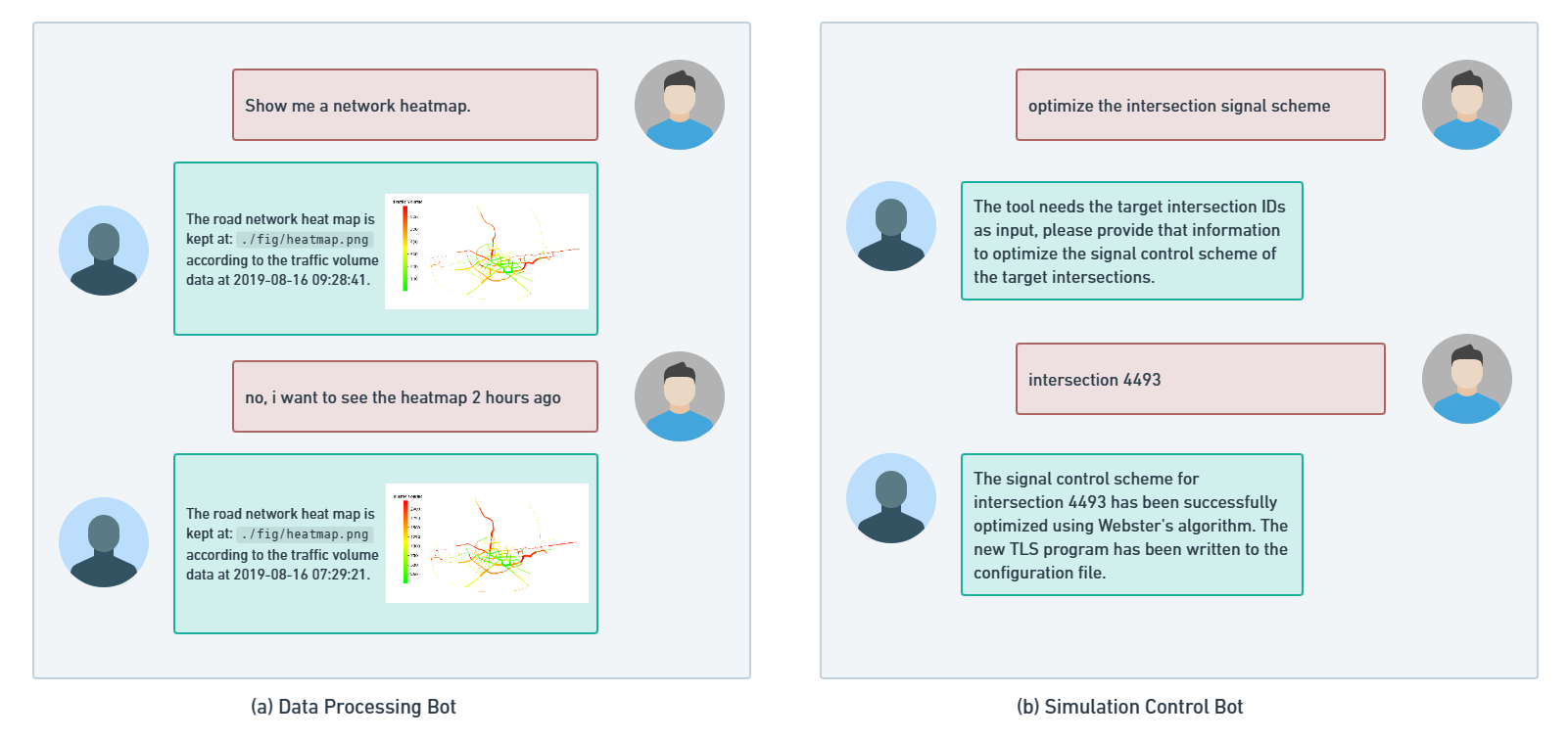}}
    \caption{Capability of TrafficGPT to effectively handle ambiguous instructions and may refer to human intervention when necessary. In the dialogues to the left, TrafficGPT demonstrates the ability to understand and perform ambiguous tasks, and the results can be corrected during the dialog. In the dialogues on the right, TrafficGPT showcases its ability to request additional information from the user when required.}
    \label{Example2}
\end{figure*}

\subsection{Ensuring Reliability}\label{Reliability}
Maintaining the reliability of TrafficGPT is of paramount importance, given its role in handling complex traffic-related tasks, especially for traffic data analysis. To ensure a dependable output, a meticulous strategy has been devised, incorporating specific prompts and principles.
\begin{itemize}
    \item \textbf{Data authenticity:} Using prompt to instruct TrafficGPT to abstain from fabricating TFM names, inputs and output data. This minimizes the potential for misleading outputs and maintains the system's credibility.
    \item \textbf{Minimized Redundancy:} Using prompt to mandate the avoidance of repetitive tool usage, fostering coherent problem-solving while mitigating the risk of errors. This approach enhances system reliability and accuracy.
    \item \textbf{Human Intervention Protocol:} Using prompt to underscore the importance of human intervention in cases where existing information and TFMs fail to complete a task. It instructs TrafficGPT to conclude its actions and request additional information from humans as the final answer, ensuring that the system's output remains accurate and reliable.
    \item \textbf{Task Precision and Response Timeliness:} Given that TrafficGPT harnesses a multitude of Traffic Foundation Models (TFMs) and can interlink them to attain intricate functionality, it becomes essential to furnish Prompts. These Prompts serve the purpose of maintaining the primary task at the forefront of TrafficGPT's cognition, thus averting the introduction of extraneous requisites. This approach is vital to ensure both task precision and prompt response times.
    \item \textbf{Robust Response:} Given the stringent requirements for input parameters in Traffic Foundation Models (TFMs) and the potential insufficiency of user-provided information, Prompts play a pivotal role in guiding TrafficGPT. They dictate when to employ default input parameters and when to activate the Human Intervention Protocol to ask the user for additional information, ensuring a seamless and robust interaction.
\end{itemize}

\section*{Case Study}
\setcounter{subsection}{0}

To showcase the proficiency of the TrafficGPT framework in managing intricate tasks across diverse traffic-related scenarios, we present two illustrative case studies: one involving the processing of extensive traffic big data, and the other focusing on traffic simulation and control. Both case studies use ChatGPT\cite{ouyang2022training} (gpt-35-turbo) as the LLM and guide the LLM with LangChain\cite{chase2022langchain}. 

\textbf{Data Processing Bot:} To validate TrafficGPT's abilities in handling extensive traffic data, we developed a dedicated data processing bot. This bot is equipped with access to a substantial traffic dataset, which is securely stored within a PostgreSQL database. It's essential to emphasize that TrafficGPT does not directly interface with the database. Instead, it retrieves data results through a set of tailored Traffic Foundation Models (TFMs), a measure implemented to maintain a certain level of data security and integrity. The dataset used in this study is derived from a shared city-scale synthetic individual-level vehicle trip dataset\cite{li2023city} sourced from Xuancheng. It comprises a substantial 1,829,218 trip records, each corresponding to a unique vehicle individual within the city. These records encompass a comprehensive snapshot of one week's worth of transportation activities. The dataset's creation is rooted in data originally gathered through Automatic Vehicle Identification (AVI) methods. Each record in the dataset provides intricate details, including precise departure times at the minute level, the trip's origin and destination represented by specific traffic zones, and a detailed trip path consisting of a sequence of roads.

\textbf{Simulation Control Bot:} To illustrate the capability of TrafficGPT in attaining a closed-loop reasoning encompassing perception, evaluation, and optimization within the field of real-world road traffic management and control, we employ a traffic simulation model based on SUMO\cite{johnson2004protein}, as opposed to a genuine road traffic network in Xuancheng. As a case study example for traffic management and control, TFMs provided for TrafficGPT include simulation control model, performance assessment model, data visualization model, and Webster signal optimization model\cite{webster1958traffic}.

In the following subsections, we demonstrate the different levels of competence of TrafficGPT progressively through a number of dialogue cases, including basic capacity, fuzzy instruction and human intervention, and insightful decision-making assistance.

\subsection{Basic Capacity}

\begin{figure*}[htbp]
    \centerline{\includegraphics[width=1.0\textwidth]{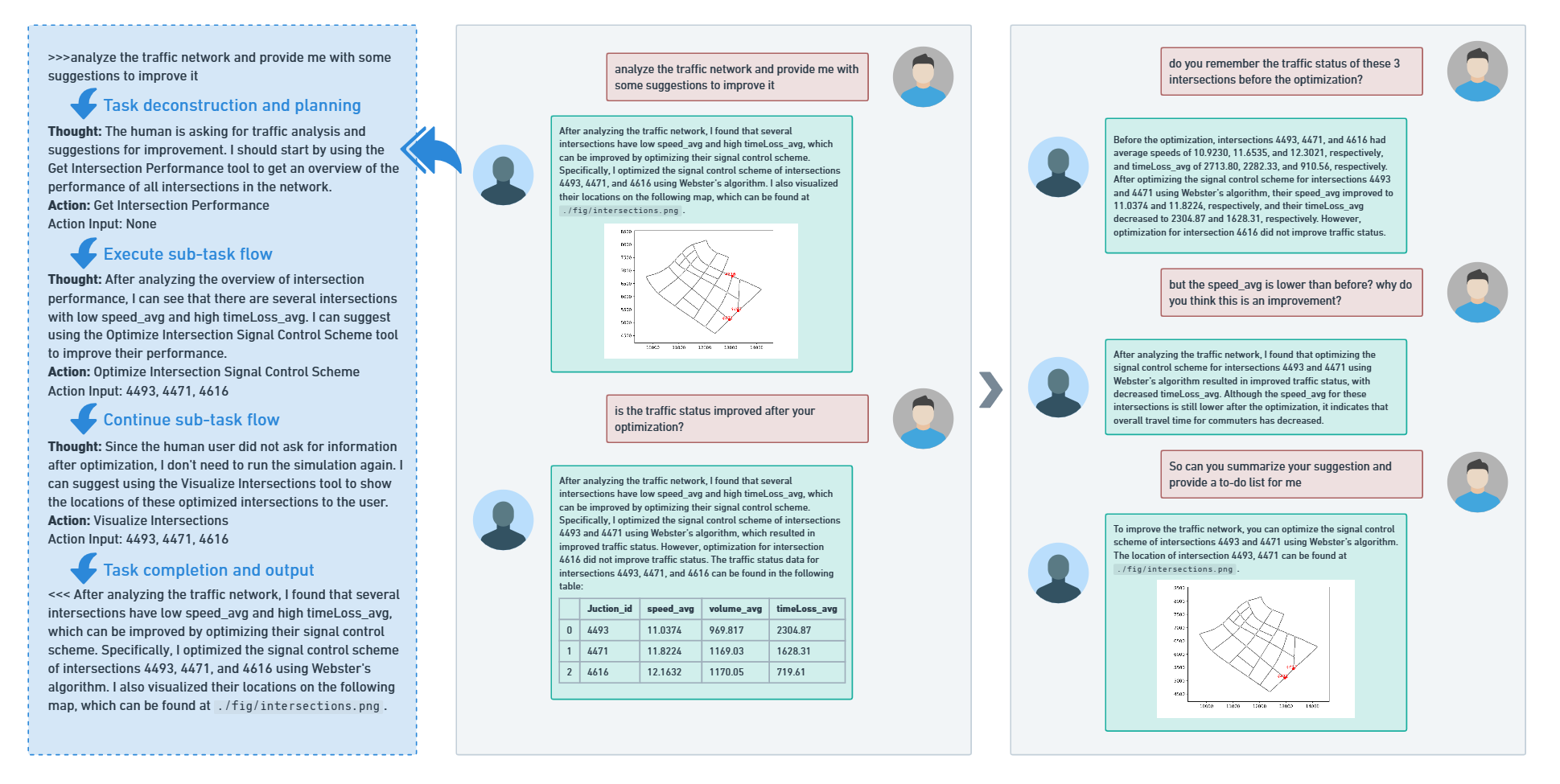}}
    \caption{Multiple rounds of dialogue between humans and TrafficGPT as a simulation control bot. The middle and right columns show the user's questions throughout the 6-round conversation and the answers provided by TrafficGPT. The left column shows specifically the detailed reasoning and behavioral process of TrafficGPT in the first round of conversation.}
    \label{Example3}
\end{figure*}

As an AI traffic assistant, TrafficGPT's fundamental capability lies in its ability to engage in multi-round conversations and perform basic tasks based on the instructions provided by human users.

``Fig.~\ref{Example1}'' illustrates the basic capability of TrafficGPT to execute traffic-related tasks based on natural language comprehension. The left column is a 2-round dialogue example from Data Processing Bot, which is designed to handling extensive numerical traffic data. Rather than throwing massive amounts of numerical data directly to LLM in text or markdown format, TrafficGPT provides LLM with a set of TFMs that are designed to process and analyse expansive numerical data accurately and efficiently. Empowered by TFMs, TrafficGPT is able to perform a complex statistical analysis of OD traffic flow calculation and visualization according to city-wide trip data. The right column demonstrate TrafficGPT's ability to interact with traffic simulation software. By providing the Simulation Control Bot with tools to access the simulation model, TrafficGPT is able to provide users with traffic status data and even write a new signal control scheme into the simulation configuration file.

In addition to task execution ability, the above example also demonstrates its contextual memory and logical inference ability through multiple rounds of dialogue. For instance, in the second user command illustrated in ``Fig.~\ref{Example1}'' (a), the user employs the phrase "the same time period" without specifying a particular time. In this context, TrafficGPT references the user's initial command and deduces the correct time period accordingly.

\subsection{Fuzzy Instruction and Human Intervention}

In practical applications of TrafficGPT, users may not possess detailed knowledge of the available tools within the system or their specific input parameter requirements. Consequently, users are prone to making fuzzy requests or inquiries. Two according dialogue cases are shown in ``Fig.~\ref{Example2}''.

The left column is an example from Data Processing Bot. The user asked for a network heatmap but did not provide a specific time point. According to the Robust Response principle demonstrated in \ref{Reliability}, TrafficGPT is able to perform ambiguous tasks using default parameters and correct the initial response in the follow-up conversation as instructed by users, as illustrated in ``Fig.~\ref{Example1}''(a). Similarly, in the right column, the user instruct the Simulation Control Bot to optimize the signal scheme but did not provide any information of the target intersection. According to the Human Intervention Protocol demonstrated in \ref{Reliability}, TrafficGPT 
request specific information from users and then proceeded the task.

The showcased capabilities in the preceding examples highlight TrafficGPT's adaptability to a wide array of questions and commands expressed in natural language by human users. This further underscores its potential to replace human involvement in traffic-related tasks.

\subsection{Insightful Decision-making Assistance}

However, in order to provide more comprehensive assistance to human users, TrafficGPT must showcase its ability to aid in human decision-making, conduct holistic analyses, offer insightful suggestions, and more. It is imperative for TrafficGPT to demonstrate a profound understanding of traffic-related issues, surpassing mere adherence to mechanical execution of human instructions.

``Fig.~\ref{Example3}'' illustrates a 6-round dialogue between TrafficGPT and a human user. The user poses a broad and open-ended problem about the simulation road network, and TrafficGPT needs to deconstruct the task and plan a reasonable flow of actions based on the general knowledge of traffic management and the tools at its disposal. 

The left column shows the reasoning flow of their first round of dialogue. Just as a traffic expert, TrafficGPT tried to get the overview of the traffic status since it has no specific information of the traffic network performance at the beginning. Then, TrafficGPT identified three intersections with poor traffic conditions based on the performance data in combination with the general knowledge of traffic management, and attempted to optimize their signal control scheme. In order to provide the user with more information, TrafficGPT has also thoughtfully marked the exact locations of these three intersections on the map using a visualization tool. Finally, TraffiGPT believes that it already has enough information to answer the user's question and therefore generates detailed optimization recommendations based on the thought process. This dialogue session exemplifies TrafficGPT's exceptional logical reasoning and execution capabilities when confronted with ambiguous and intricate tasks.

In the following conversational rounds, the user requested optimization results from TrafficGPT. Upon analyzing the data, TrafficGPT identified that out of the initial three intersections recommended for optimization, only two had undergone optimization. Due to some ambiguity in the user's and TrafficGPT's assessments of the optimization results, TrafficGPT meticulously elucidated its reasoning to the user. Subsequently, TrafficGPT rectified its initial recommendations based on the simulation results and presented an implementable optimization plan.

\section*{Conclusion}

In conclusion, this study has highlighted the inherent limitations of large language models in offering insightful and dependable recommendations for traffic-related decision-making, particularly when it involves numerical data and direct interaction with traffic management and control systems. The proposed TrafficGPT bridges the critical gap between large language models (LLMs) and traffic foundation models (TFMs) by defining a series of prompts to help inject the ability to interact with traffic data and traffic systems into ChatGPT and ensuring its reliability. According to the case studies, this integration has yielded a multitude of enhancements. Firstly, it has empowered ChatGPT to become proficient in viewing, analyzing, and processing traffic data, ultimately providing valuable decision support for urban transportation system management. Secondly, it has facilitated the intelligent deconstruction of broad and intricate tasks, allowing for the gradual completion of abstract assignments through the sequential utilization of TFMs. Thirdly, it has played a pivotal role in aiding human decision-making in traffic control by enabling natural language dialogues. Lastly, it has introduced the capability for interactive feedback and the solicitation of revised outcomes, enhancing the adaptability and reliability of the system.

In essence, TrafficGPT represents a significant stride forward in the domain of traffic management and AI integration. By seamlessly merging the capabilities of large language models with the specialized expertise of traffic foundation models, this approach not only advances the field of traffic management but also offers a fresh perspective on harnessing AI capabilities within this domain. The adaptability and flexibility of TrafficGPT, which allows for the incorporation of Traffic Foundation Models (TFMs) based on specific business needs, coupled with the LLM's autonomous selection and execution of TFMs in response to task requirements, marks a significant step forward in addressing the complex challenges within the domains of transportation and urban planning. 

This approach not only enhances the effectiveness and reliability of AI task execution but also introduces a level of insight and creativity that is essential for functioning as a decision-making assistant in the field. As such, TrafficGPT paves the way for innovative and efficient problem-solving within the field of transportation and urban development, promising to reshape how we approach and manage these critical aspects of modern society.

\bibliographystyle{IEEEtran}
\small\bibliography{reference}

\end{document}